\pdfoutput=1

\documentclass[11pt]{article}

\usepackage{naacl2021}

\usepackage{times}
\usepackage{latexsym}

\usepackage[T1]{fontenc}

\usepackage[utf8]{inputenc}

\usepackage{microtype}
\usepackage{amsthm}
\usepackage{amsfonts}
\usepackage{amsmath}
\usepackage{amssymb}
\usepackage{tikz}
\usepackage{tikz-cd}
\usepackage[ruled,vlined]{algorithm2e}
\usepackage{algpseudocode}
\usepackage{comment}
\usepackage{paralist} 
\usepackage{cleveref}

\title{Towards a Model-Theoretic View of Narratives}

\let\OldTexttt\texttt
\renewcommand{\texttt}[1]{\OldTexttt{\small{#1}}}

\newtheorem{theorem}{Theorem}[section]

\theoremstyle{definition}
\newtheorem{definition}{Definition}[section]

\theoremstyle{definition}

\author{Louis Castricato\footnote{equal contribution} \\
  Georgia Tech \\
  EleutherAI\\
  \texttt{lcastric@gatech.edu} \\\And
  Stella Biderman\footnotemark[1] \\
  Georgia Tech\\
  EleutherAI\\
  \texttt{stella@eleuther.ai} \\\And
  Rogelio E. Cardona-Rivera \\
  University of Utah\\
  \texttt{rogelio@cs.utah.edu} \\\And
  David Thue\\
  Carleton University \\
  \texttt{david.thue@carleton.ca} \\
  }

\begin{document}
\maketitle
\begin{abstract}

In this paper, we propose the beginnings of a formal framework for modeling narrative \textit{qua} narrative. Our framework affords the ability to discuss key qualities of stories and their communication, including the flow of information from a Narrator to a Reader, the evolution of a Reader's story model over time, and Reader uncertainty. We demonstrate its applicability to computational narratology by giving explicit algorithms for measuring the accuracy with which information was conveyed to the Reader and two novel measurements of story coherence.

\end{abstract}

\section{Introduction}

Story understanding is both %
\begin{inparaenum}[(1)]
\item the process through which a cognitive agent (human or artificial) mentally constructs a \emph{plot} through the perception of a \emph{narrated discourse}, and
\item the outcome of that process: \textit{i.e.}, the agent's mental representation of the plot.
\end{inparaenum}
The best way to computationally model story understanding is contextual to the aims of a given research program, and today we enjoy a plethora of artificial intelligence (AI)-based capabilities.

{Data-driven} approaches---including statistical, neural, and neuro-symbolic ones---look to narrative as a benchmark task for demonstrating human-level competency on {inferencing},  {question-answering}, and {storytelling}. That is, they draw associations between event~\citep{chambers2008unsupervised}, causal~\citep{li2012crowdsourcing}, and purposive~\citep{jiang2018learning} information extracted from textual or visual narrative corpora to answer questions or generate meaningful stories that depend on information implied and not necessarily expressed by stories~\citep[{e.g.}][]{roemmele2011choice,mostafazadeh2016corpus,martin2018event,kim2019progressive}.

{Symbolic} approaches seek to understand narrative, its communication, and its effect by using AI techniques as computational modeling tools, including logic, constraint satisfaction, and automated planning. These include efforts to model creative storytelling as a search process~\citep{riedl2006story,ThueICIDS2016}, generating stories with predictable effects on their comprehension by audiences~\citep{cardona-rivera2016question}, and modeling story understanding through human-constrained techniques~\citep{martens2020vne}.

However, despite excellent advances, few works have offered a thorough conceptual account of narrative in a way that affords reconciling how different research programs might relate to each other.
%
%
Without a foundation for shared progress, our community might strain to determine how individual results may build upon each other to make progress on story understanding AI that performs as robustly and flexibly as humans do~\citep{cardona-rivera2019desiderata}.
In this paper, we take steps toward such a foundation.

We posit that such a foundation must acknowledge the diverse factors that contribute to an artifact being treated \emph{as} a narrative. Key among these factors is a narrative's \emph{communicative} status: unlike more-general natural language generation~\citep[cf.][]{gatt2018survey}, an audience's \emph{belief dynamics}---the trajectory of belief expansions, contractions, and revisions~\citep{alchourron1985logic}---is core to what gives a narrative experience its quality~\citep{herman2013storytelling}. Failure to engage with narratives on these grounds risks losing an essential aspect of what makes narrative storytelling a vibrant and unique form of literature.

To that end, we define a preliminary theoretical framework of narrative centered on information entropy. Our framework is built atop \emph{model theory}, the set-theoretic study of language interpretation. Model theory is a field of formal logic that has been used extensively by epistomologists, linguists, and other theorists as a framework for building logical semantics. 

\paragraph{Contributions} In this paper, we propose the beginnings of a formal framework for modeling narrative \textit{qua} narrative. Our framework includes the ability to discuss the flow of information from a Narrator to a Reader, the evolution of a Reader's story model over time, and Reader uncertainty. Our work is grounded in the long history of narratology, drawing on the rich linguistic and philosophical history of the field to justify our notions.

We use our framework to make experimentally verifiable conjectures about how story readers respond to under-specification of the story world and how to use entropy to identify plot points. We additionally demonstrate its applicability to computational narratology by giving explicit algorithms for measuring the accuracy with which information was conveyed to the Reader and two novel measurements of story coherence.

\section{Pre-Rigorous Notions of Narrative}\label{sec:pre-rigorous}
Before we can begin to define narrative in a formal sense, we must  examine the intuitive notions of what narrative \textit{is supposed to mean}. While we cannot address all of the complexity of narratology in this work, we cover key perspectives.

\subsection{Narratives as Physical Artifacts}\label{sec:physical}

We begin with the structuralist account within narratology; it frames a \emph{narrative} (\emph{story}) as a communicative, designed artifact---the product of a \emph{narration}, itself a realization (e.g.~book, film) of a \emph{discourse}~\citep{huhn2013narration}. The discourse is the story's information layer~\citep{genette1980narrative}: an author-structured, temporally-organized subset of the \emph{fabula}; a discourse projects a fabula's information. The fabula is the story's world, which includes its \emph{characters}, or intention-driven agents; \emph{locations}, or spatial context; and \emph{events}, the causally-, purposely-, and chronologically-related situation changes~\citep{bal1997narratology,rimmon-kenan2002narrative}.

As a designed artifact, a narrative reflects \emph{authorial intent}.  Authors design the stories they tell to affect audiences in specific ways; their designs ultimately target effecting change in the minds of audiences~\citep{bordwell1989making}. This design stems from the authors' understanding of their fabula and of the narration that conveys its discourse. When audiences encounter the designed artifact, they perform story understanding: they attempt to mentally construct a  fabula through the perception of the story's narration. 

\subsection{Narratives as Mental Artifacts}\label{sec:mental}

Story psychologists frame the narration as instructions that guide story understanding~\citep{gernsbacher1990investigation}. The fabula in the audience's mind is termed the \emph{situation model}---a mental representation of the virtual world and the events that have transpired within it, formed from information both explicitly-narrated and inferable-from a narration~\citep{zwaan1998situation}. The situation model itself \emph{is} the audience's understanding; it reflects a tacit belief about the fabula, and is manipulated via three (fabula-belief) update operations. These work across memory retrieval, inferencing, and question-answering cognition:
\begin{inparaenum}[(1)]
\item \emph{expansion}, when the audience begins to believe something, 
\item \emph{contraction}, when the audience ceases to believe something, and
\item \emph{revision}, when the audience expands their belief and contracts newly inconsistent beliefs.
\end{inparaenum}

\subsection{Narratives as Received Artifacts}\label{sec:recieved}


To the post-structuralist, the emphasis that the psychological account puts on the author is fundamentally misplaced \citep{death}. From this point of view, books are meant to be \textit{read}, not written, and how they influence and are interpreted by their readers is as essential to their essence as the intention of the author. In ``Death of the Author'' Barthes \citep{death} reinforces this concept by persistently referring to the writer of a narrative not as its creator or its author, but as its sculptor - one who shapes and guides the work but does not dictate to their audience its meaning.

\section{A Model Theoretic View of Narrative} \label{sec:model-theory}
The core of our framework for modeling narrative come from a field of mathematical logic known as model theory. Model theory is a powerful yet flexible framework that has been a heavily influential on people working in computer science, literary theory, linguistics, and philosophy \citep{sider}. Despite the centrality of model theory in our framework, a deep understanding of the topic is not necessary to work with it on an applied level. Our goal in this section is thus to give an intuitive picture of model theory that is sufficient to understand how we will use it to talk about narratives. We refer an interested reader to \citet{sider, changk} for a more complete presentation of the subject.

\subsection{An Outline of Model Theory}\label{sec:model-outline}

The central object of study in model theory is a ``model.'' Loosely speaking, a model is a world in which particular propositions are true. A model has two components: a domain, which is the set of objects the model makes claims about, and a theory, which is a set of consistent sentences that make claims about elements of the domain. Models in many ways resemble fabulas, in that they describe the relational properties of objects. 
Model theory, however, requires that the theory of a model be \textit{complete} – every expressible proposition must be either true or false in a particular model. 

Meanwhile, our notion of a fabula can be \textit{incomplete} - it can leave the truth of some propositions undefined. 
This means that the descriptions we are interested in do not correspond to only \textit{one} model, but rather that there is an infinite set of models that are consistent with the description. This may seem limiting, but we will show in \Cref{sec:plausibility} that it is actually amenable to analysis.

As an example, consider a simple world in which people can play cards with one another and wear clothes of various colours. The description \textit{``Jay wears blue. Ali plays cards with Jay.''} is incomplete because it does not say what colours Ali wears nor what other colours Jay wears. This description is consistent with a world in which there are characters other than Jay and Ali or colours other than blue (varying the domain), as well as one where additional propositions such as \textit{``Ali wears blue.''} hold (varying the theory).

Although we learn more about the domain and the theory of the narrator's model as the story goes on, we will never learn every single detail. Some of these details may not even be known to the narrator! For this reason, our framework puts a strong emphasis on consistency between models, and on the set of all models that are consistent with a particular set of statements.

Another very important aspect of model theory is that it is highly modular. Much of model theory is independent of the underlying logical semantics, which allows us to paint a very general picture. If a particular application requires augmenting the storytelling semantics with additional logical operators or relations, that is entirely non-problematic. For example, it is common for fabulas to contain Cause(X, Y) := ``X causes Y'' and Aft(X, Y) := ``Y occurs after X.'' Although we don't specifically define either of these relations, they can be included in a particular application by simply adding them to the underlying logic.

\subsection{Story-World Models and the Fabula}

As detailed in \cref{sec:pre-rigorous}, the fabula and story-world (i.e.~situation) model are two central components of how people talk about storytelling. In this section we introduce formal definitions of these concepts and some of their properties.

\begin{definition}
    A language, $\mathcal{L}$, is a set of rules for forming syntactically valid propositions. In this work we will make very light assumptions about $\mathcal{L}$ and leave its design largely up to the application.
\end{definition}

A language describes \textit{syntactic} validity, but doesn't contain a notion of truth. For that, we need a model.

\begin{definition}
    A story world model, $S$, over a language $\mathcal{L}$ is comprised of two parts: a domain, which is the set of things that exist in the story, and an interpretation function, which takes logical formulae and maps them to corresponding objects in the domain. In other words, the interpretation function is what connects the \textit{logical expression} ``A causes B'' to the signified \textit{fact in the world} that the thing we refer to as A causes the thing we refer to as B.
\end{definition}
\begin{definition}
    The theory of a story world model, $S$, is the set of all propositions that are true in $S$. It is denoted $\tilde{S}$. When we say ``$P$ is true in the model $S$'' we mean that $P\in S'$.
\end{definition}

Formalizing the concept of a fabula is a bit trickier. Traditionally, fabulas are represented diagrammatically as directed graphs. However this representation gives little insight into their core attributes. We posit that, at their core, fabulas are relational objects. Specifically, they are a collection of elements of the domain of the story-world model together with claims about the relationships between those objects. Additionally, there is a sense in which the fabula is a ``scratch pad'' for the story-world model. While a reader may not even be able to hold an entire infinite story-world model in their head, they can more easily grasp the distillation of that story-world model into a fabula.

\begin{definition}
    A reasoner's fabula for a story world model $S$, denoted $F$, is a set of propositions that makes claims about $S$. A proposition $P$ is a member of $F$ if it is an explicit belief of the reasoner about the narrative that the reasoner deems important to constructing an accurate story-world model.
\end{definition}

\section{Conveying Story Information}

An important aspect of stories is that they are a way to convey information. In this section, we will discuss how to formalize this process and what we can learn about it. Although stories can be constructed and conveyed in many different ways, we will speak of a Narrator who tells the story and a Reader who receives it for simplicity.

\begin{figure}[ht]
    \centering
    \begin{tikzcd}
S_N \arrow[dd, "\phi" description]\arrow[rr, "d'" description, dashed]     &           & S_R\\
                                                                    &           &                               \\
F_N \arrow[rr, "d" description]                                  &           & F_R \arrow[uu, "\psi" description]                         
\end{tikzcd}
    \caption{Commutative diagram outlining storytelling}
    \label{fig:retelling_1}
\end{figure}
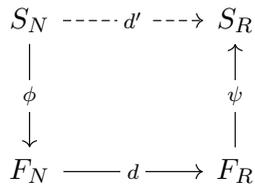

The core of our model of storytelling as an act of communication can be seen in Figure \ref{fig:retelling_1}.

This diagram represents the transmission of information from the Narrator's story-world to the Reader's, with each arrow representing the transmission from one representation to another. In an idealized world, stories would be conveyed by $d'$: straight from the story world of the narrator ($S_N$) to the story world of the reader ($S_R$). In actuality, narrators must convey their ideas through media\footnote{Nevertheless, having a conception of $d'$ is very important on a formal level as we will see later.}. To do this, the narrator compresses their mental story world (via $\phi$) into a fabula ($F_N$) which is then conveyed to the reader via speech, writing, etc. The conveyance of the fabula \textit{as understood by the Narrator} ($F_N$) to the fabula \textit{as understood by the Reader} ($F_R$) is denoted in our diagram by $d$. $d$ is in many ways the real-world replacement for the function $d'$ the Narrator is unable to carry out. Once the discourse has been consumed by the Reader, the Reader then takes their reconstructed fabula ($F_R$) and uses the received information to update their story world model ($S_R$, via $\psi$).

\subsection{Accurately Conveying Information}

Often times, information conveyed from the Narrator to the Reader is ``conveyed correctly.'' By this, we mean that the essential character of the story was conveyed from the Narrator to the Reader in such a way that the Reader forms accurate beliefs about the story-world. While accuracy is not always a primary consideration - some stories feature unreliable narrators or deliberately mislead the Reader to induce experiences such as suspense, fear, and anticipation - the ability to discuss the accuracy and consistency of the telling of the story is an essential part of analyzing a narrative.

The $d'$ arrow in our diagram suggests a reasonable criteria for accurate conveyance: a story is accurately conveyed if the path $S_N\to F_N\to F_R\to S_R$ and the path $S_N\dashrightarrow S_R$ compute the same (or, in practice, similar) functions. In mathematics, this property of path-independence is known as commutativity and the diagram is called a ``commutative diagram'' when it holds. For the purposes of narrative work, the essential aspect is that the arrows ``map corresponding objects correspondingly.'' That is, if a story is accurately conveyed from $N$ to $R$ then for each proposition $P\in S_N$ there should be a corresponding $P'\in S_R$ such that the interpretations of $P$ and $P'$ (with respect to their respective models) have the same truth value and $(\phi\circ d\circ\psi)(P) = P'$. In other words, $P$ and $P'$ make the same claims about the same things.

\subsection{Time-Evolution of Story-World Models}

The transference of information depicted in \cref{fig:retelling_1} gives rise to a straightforward way to understand how the Reader gains knowledge during the course of the story and incorporates new information into their existing story-world model. One pass through the diagram from $S_N$ to $S_R$ represents ``one time step'' of the evolution of the Reader's world model\footnote{For simplicity we will speak of this as a discrete time series, though for some media such as film it may make sense to model it as a continuous phenomenon.}.

Iterating this process over the the entire work gives a time series of story-world models, $S_R(t)$, with $S_R(i)$ representing the Reader's story-world model at time $t=i$. We are also typically interested in how the story-world model changes over time, as the Reader revises their understanding of the story-world through consuming the discourse. This will be the subject of the next section.

\section{A Detailed Look at Temporal Evolution, with Applications to Plot}\label{sec:temporal}

A common accepted notion in narratology literature is that at any given moment a reader contains a potentially infinite set of possible worlds. Determining which of these worlds agree with each other is a required attribute for consuming discourse. How do we discuss the notion of collapsing possible worlds upon acquiring new knowledge?

Assume that we have a narrator, $N$, and reader $R$ with fabulas $F_N$ and $F_R$ respectively. Given our definition of a story world model, $S$, we define $\mathbf{S}(t)$ as the set of all world models that satisfy $F_R(t)$. Let $\mathbf{\rho}_{t+1}$ refer to the set of formulae that are contained in $F_R(t + 1)\backslash F_R(t)$. Let
\[S'_R(t+1) = \mathbf{S}_R(t + 1) \cap \mathbf{S}_R(t)\]
and similarly
\[\tilde{S}'_R(t+1) = \tilde{\mathbf{S}}_R(t + 1) \cap \tilde{\mathbf{S}}_R(t)\] refer to the shared world models between the two adjacent time steps. Note that it must follow $\forall \rho \in \mathcal{P}_{t+1}$, $\forall s \in \tilde{\mathbf{S}}'_R(t+1)$, $\rho \in s$. That is to say, the story worlds that remain between the two time steps are the ones that agree on the propositions added by consuming $F_N(t+1)$. Since this can be repeated inductively, we can assume that for any such $t$ we have that all such models agree on all such provided propositions.

Something to note that for $\rho \in \mathcal{P}_{t+1}$, $\rho$ will always be either true or false in $\tilde{S}_R(t)$- regardless if it is expressed in the fabula or not since $\tilde{S}_R(t)$ is the logical closure of $S_R(t)$.

\subsection{Collapse of Worlds over Time}

Something to note is that a set of story worlds $\tilde{\mathbf{S}}_R(t)$ does not provide us a transition function to discuss how the world evolves over time. Furthermore, there is no reasonable way to infer $\tilde{S}_R(t) \mapsto \tilde{S}_R(t + 1)$, as $\tilde{S}_R(t)$ provides no information about the actions that could inhibit or allow for this transition- it simply provides us information about if a proposition is true within our story world. To rectify this, we need to expand our commutative diagram to act cross-temporally. The full diagram can be found in the appendix.

Let $\zeta_N$ denote the transition function from $F_N(t)$ to $F_N(t + 1)$. Define $\zeta_R$ likewise. See Figure \ref{fig:zeta_1} on page \pageref{fig:zeta_1}. Note that there is no inherent general form of $\zeta_N$ or $\zeta_R$ as they are significantly context dependent. One can think of them as performing graph edits on $F_N$ and $F_R$ respectively, to add the new information expressed in $S_N(t+1)$ for $\zeta_N$ and $(d \circ \phi)(S_N(t+1))$ for $\zeta_R$. 

The objective of $\zeta_R$ in turn is to guide the fabula to reach goals. This imposes a duality of $\psi$ and $\zeta_R$. $\psi$ attempts to generate the best candidate story worlds for the reader's current understanding, where as $\zeta_R$ eliminates them by the direction the author wants to go.

This in turn brings us to the notion of compression and expansion. Namely that $\psi$, if left unchecked, will continuously expand the fabula. In turn $\zeta_R$ is given the goal of compressing the story worlds that $\psi$ produces by looking at the resulting transition functions that best match the author's intent.\footnote{There is no single best way to define an author's intent. For instance, we could have easily said that $\psi$ denotes author intent while $\zeta_R$ determines which intents are best grounded in reality. The choice, however, needs to be made.}

\subsection{Plot Relevance}

Stories contain many different threads and facts, and it would be nice to be able to identify the ones that are relevant to the plot. We begin with the idea of the relevance of one question to another.

\begin{definition}
Consider a question about a story, $q$, of the form ``if A then B" with possible values for $A=\{T, F\}$ and possible values for $B=\{T,F\}$. We say that the {\bf relevance} of $B$ to $A$ given some prior $\gamma$ is 
    \begin{equation}
    H(A = a_i | \gamma) -  H(B = b_j | A = a_i, \gamma) 
    \end{equation}
    where $a_i$ and $b_j$ are the true answers to $A$ and $B$ and $H$ refers to binary entropy.
\end{definition}

Note that the relevance of $B$ to $A$ \textit{depends on the true answers}. This is perhaps surprising, but after some consideration it should be clear that this has to be true. After all, the causal relationship between $A$ and $B$ could depend on the true answers! Consider the case where $A$ is ``is Harry Potter the prophesied Heir of Slytherin?'' and $B$ is ``can Harry Potter speak Parseltongue because he is a descendent of Slytherin?'' If Harry is a blood descendant of Slytherin and that's why he can speak Parseltongue, then $B$ is highly relevant to $A$. However, the actual truth of the matter is that Harry's abilities are completely independent of his heritage and arose due to a childhood experience. Therefore $B$ does not in fact have relevance to $A$ even though \textit{it could have had relevance} to $A$.

Having defined a notion of the relevance of Question $A$ to Question $B$, our next step is connecting to existing narratological analysis. Consider Barthes' notion of kernels and satellites.\cite{barthes1975introduction}

\begin{definition}
    A \textbf{kernel} is a narrative event such that after its completion, the beliefs a reader holds as they pertain to the story have drastically changed.\footnote{The notion of "drastic" is equivalent to "majority." To rigoriously define Barthes' Kernel, and hence Barthes' Cardinal, we would require ultraproducts- which is outside of the scope of this paper.}
\end{definition}

\begin{definition}
    A \textbf{satellite} is a narrative event that supports a kernel. They are the minor plot points that lead up to major plot points. They do not result in massive shift in beliefs.
\end{definition}

Of importance to note is that satellites imply the existence of kernels, e.g. small plot points will explain and lead up to a large plot point, but kernels do not imply the existence of satellites- kernels do not require satellites to exist. One can think of this as when satellites exist kernels must always exist on their boundary whether they are referred to in the text or not.

A set of satellites, $s = \{s_1, \hdots, s_n\}$, is said to be relevant to a kernel, $k$, if after the kernel's competition, the reader believes that the set of questions posed by $k$ are relevant to their understanding of the story world given prior $s$.
dh
Take note of the definition of relevance. Simply put, $A$ denotes the questions that define some notion of story world level coherency where as $B$ denotes the set of questions that define some notion of transitional coherency.

\section{Possible Worlds and Reader Uncertainty}\label{sec:plausibility}

So far we have spoken about the Reader's story-world model as if there is only one, but in light of the discussion in \cref{sec:model-theory} it is unclear it truly makes sense to do so. In actuality, the Reader never learns to ``true story-world model'' (insofar as one can even be said to exist). Rather, the Reader has an evolving set of ``plausible story-world models'' that are extrapolated based on the incomplete information conveyed in the story. The purpose of this section is to detail how these ``plausibilities'' interact with each other and with plausibilities at other time steps.

It likely seems natural to model the Reader's uncertainty with a probabilistic model. Unfortunately, the topological structure of first-order logic makes that impossible as there is no way to define a probability distribution over the set of models that are consistent with a set of sentences. Instead, we are forced to appeal to \textit{filters}, a weaker notion of size that captures the difference between ``large'' and ``small'' sets. Again we develop the theory of ultrafilters only to the extent that we require, and refer an interested reader to a graduate text in mathematical logic for a thorough discussion.

\begin{definition}
    Let $Q$ be a set of sentences that make claims about a narrative. A non-empty collection $\mathcal{F}_w\subseteq\mathcal{P}(Q)$ is a weak filter iff
    \begin{enumerate}
        \item $\forall X, Y\in\mathcal{P}(Q),\;X\in \mathcal{F}_w$ and $X\subseteq Y\subseteq\mathcal{P}(Q)$ implies $Y\in\mathcal{F}_w$
        \item $\forall X\in\mathcal{P}(Q),\;X\not\in \mathcal{F}_w$ or $\mathcal{P}(Q)\backslash X\not\in \mathcal{F}_w$
    \end{enumerate}
\end{definition}

We say that $\mathcal{F}_w$ is a weak ultrafilter and denote it $\mathcal{UF}_w$ if the second requirement is replaced by $\forall X\in\mathcal{P}(Q),\; X\in \mathcal{F}_w\iff \mathcal{P}(Q)\backslash X\not\in \mathcal{F}_w$ \cite{kmp}.

A reader's beliefs at time $t$ defines a weak filter over the set of possible story-world models $\{S^i_R\}$. Call this filter $\mathcal{F}_w$, dropping the $t$ when it is clear from context. Each element $U\in\mathcal{F}_w$ is a set of story world models that define a \textit{plausibility}. This plausibility describes a set of propositions about the story that the reader thinks paints a coherent and plausible picture. Formally, a plausibility identified with the largest set of sentences that is true for every model in $U$, or $\cap_{S\in U} T(S)$ where $T(S)$ denotes the set of true statements in $S$. That is, the set of plausible facts.

The intuition for the formal definition of a weak filter is that 1. means that adding worlds to an element of the filter (which decreases the number of elements in $\cap_{S\in U} T(S)$) doesn't stop it from describing a plausibility since it is specifying fewer facts; and that 2. means that it is not the case that both $P$ and $\neg P$ are plausible. It's important to remember that membership in $\mathcal{F}_w$ is a binary property, and so a statement is either plausible or is not plausible. We do not have shades of plausibility due to the aforementioned lack of a probability distribution.

As a framework for modeling the Reader's uncertainty, weak filters underspecify the space of plausible story world as a whole in favor of capturing what the reader ``has actively in mind'' when reading. This is precisely because the ultrafilter axiom is not required, and so for some propositions neither $P$ nor $\neg P$ are judged to be plausible. When asked to stop and consider the truth of a specific proposition, the reader is confronted with the fact that there are many ways that they can precisify their world models. How a Reader responds to this confrontation is an experimental question that we leave to future work, but we conjecture that with sufficient time and motivation a Reader will build a weak ultrafilter $\mathcal{UF}_w$ that extends $\mathcal{F}_w$ and takes a position on the plausibility of all statements in the logical closure of their knowledge.

Once the Reader has fleshed out the space of plausibilities, we can use $\mathcal{UF}_w$ to build the \textit{ultraproduct} of the Reader's story-world models. An ultraproduct \citep{changk} is a way of using an ultrafilter to engage in \textit{reconciliation} and build a single consistent story world-model out of a space of plausibilities. Intuitively, an ultraproduct can be thought of as a vote between the various models on the truth of individual propositions. A proposition is considered to be true in the ultraproduct if and only if the set of models in which it is true is an element of the ultrafilter. We conjecture that real-world rational agents with uncertain beliefs find the ultraproduct of their world models to be a reasonable reconciliation of their beliefs and that idealized perfectly rational agents will provably gravitate towards the ultraproduct as the correct reconciliation.

\section{Applications to Computational Narratology}\label{sec:applications}

Finally, demonstrate that our highly abstract framework is of practical use by using it to derive explicit computational tools of use to computational narratologists.

\subsection{Entropy of World Coherence}

Firstly it is important to acknowledge that a reader can never reason over an infinite set of worlds. Therefore, it is often best to consider a finite sample of worlds. Given the (non-finite) set of story worlds, $\mathbf{S}(t)$, there must exist a set $s' \subset \mathcal{UF}_w(t)$ such that every element in $s'$ is one of the "more likely" interpretations of the story world. This notion of more likely is out of scope of this paper; however, in practice more likely simply denotes probability conditioned from $\tilde{\mathbf{S}}(t - 1)$.

It is equally important to note that every element of $s'$, by definition, can be represented in the reader's mind by the same fabula, say $F(t)$. Let $Q$ be some set of implications that we would like to determine the truth assignment of. Let $P_{s'}(q)$ refer to the proportion of story worlds in $s'$ such that $q$ is true.\footnote{An equivalent form of $P(q)$ exists for when we do not have a form of measure. Particularly, define $P(q) = 1$ when $q$ is true in the majority of story worlds, as defined by our ultrafilter. Similarly, let $P(q) = 0$ otherwise. For those with prior model theory experience, $P(q) = 1$ if $q$ holds in an ultraproduct of story world models.} Clearly, $P_{s'}(q)$ is conditioned on $s'$. We can express the entropy of this as
\begin{align*}
 &H(P_{s'}(q)) = H(q | s')\\
 &= H(A = T | s') - H(B = b_j | A = T, s')
\end{align*}
Therefore averaging over $H(P_{s'}(q))$ for all $q \in Q$ is equivalent to determining the relevance of our implication to our hypothesis. This now brings us to EWC, or entropy of world coherence. These implications are of the form ``Given something in the ground truth that all story worlds believe, then X" where X is a proposition held by the majority of story worlds but not all. We define EWC as $$\text{EWC}(s', Q) = \frac{1}{|Q|}\sum_{q \in Q} P_{s'}(q)$$

\subsection{Entropy of Transitional Coherence} 

Note our definition of plot relevance. It is particularly of value to not only measure the coherency of the rules that govern our story world but also to measure the coherency of the transitions that govern it over time. We can define a similar notion to EWC, called Entropy of Transitional Coherence, which aims to measure the agreement of how beliefs change over time. In doing so, we can accurately measure the reader's understanding of the laws that govern the dynamics of the story world rather than just the relationships that exist in a static frame. 

To understand ETC we must first delve into the dynamics of modal logic. Note that for a proposition to be ``necessary'' in one frame of a narrative, it must have been plausible in a prior frame. \cite{sider} Things that are necessary, the reader knows; hence, the set of necessary propositions is a subset of a prior frame's possible propositions. 

We must define a boolean lattice to continue

\begin{definition}
    A \textbf{boolean lattice} of a set of propositions, Q, is a graph whose vertices are elements of $Q$ and for any two $a,b \in Q$ if $a \implies b$ then there exists an edge $(a,b)$ unless $a = b$
\end{definition}

Something to note is that a boolean lattice is a directed acyclic graph (DAG) and as such as source vertices with no parents. In the case of boolean lattices, a source vertex refers to an axiom, as sources are not provable by other sources.

We define one reader at two times, denoted $\mathcal{UF}_w(t)$ and $\mathcal{UF}_w(t')$ where $t' < t$. We define a filtration of possible worlds $s'(t')$ similar to how we did in the previous section. 

Given $W(t) \in \mathcal{UF}_w(t)$, a ground truth at time $t$, we restrict our view of $W(t)$ to the maximal PW of time $t'$. This can be done by looking at $$W' = \text{argmax}_{W(t) \cap s'_i} |B(W(t)) \cap (\cap_{s \in s'_i} B(s))|$$ Reason being is that it does not make sense to query about propositions that are undefined in prior frames. This effectively can be viewed as a pull-back through the commutative diagram outlined previously. See Figure \ref{fig:zeta_1} on page \pageref{fig:zeta_1}. Something to note however is that this pullback is not necessary for ETC in the theoretical setting, as all world models would agree on any proposition not contained in their respective Boolean lattices- this is not the case when testing on human subjects. Human subjects would be more likely to guess if they are presented with a query that has no relevance to their current understanding. \cite{trabasso1982causal, mandler1977remembrance}

We can however similarly define ETC by utilizing $W'$ as our ground truth with EWC. Since $W'$ is not the minimal ground truth for a particular frame, it encodes information about the ground truth where the narrative will be going by frame $t$. Therefore, define $Q$ similarly over time $t'$ relative to $W'$. We can also use this to define $P_{s'(t')}(q)$ $\forall q \in Q$. We denote ETC as
$$\text{ETC}(s'(t'), Q) = \frac{1}{|Q|}\sum_{q \in Q} P_{s'(t')}(q)$$

ETC differs from EWC in the form of implications that reside in $Q$. Particularly since ETC wants to measure the coherency of a reader's internal transition model, $\forall q \in Q$ where $q := A \implies B$ we have that $A$ is the belief a reader holds before a kernel and that $B$ is a belief the reader holds after a kernel. Since the kernel is defined as a plot point which changes the majority of a reader's beliefs, we are in turn measuring some notion of faithfulness of $\zeta_R$.

\section*{Conclusions and Future Work}

In this paper, we defined a preliminary theoretical framework of narrative that affords new precision to common narratological concepts, including fabulas, story worlds, the conveyance of information from Narrator to Reader, and the way that the Reader's active beliefs about the story can update as they receive that information.

Thanks to this precision, we were able to define a rigorous and measurable notion of plot relevance, which we used to formalize Barthes' notions of kernels and satellites. We also give a novel formulation and analysis of Reader uncertainty, and form experimentally verifiable conjectures on the basis of our theories. We further demonstrated the value of our framework by formalizing two new narrative-focused measures: Entropy of World Coherence and Entropy of Transitional Coherence, which measure the agreement of story world models frames and faithfulness of $\zeta_R$ respectively.

Our framework also opens up new avenues for future research in narratology and related fields. While we were unable to explore their consequences within the scope of this paper, the formulation of narratives via model theory opens the door to leveraging the extensive theoretical work that's been done on models to narratology. The analysis of the temporal evolution of models in \cref{sec:temporal} suggests connections with reinforcement learning for natural language understanding. In \cref{sec:plausibility} we make testable conjectures about the behavior of Reader agents and in \cref{sec:applications} we describe how to convert our theoretical musings into practical metrics for measuring consistency and coherency of stories.

\bibliography{custom}
\bibliographystyle{acl_natbib}

\appendix
\begin{figure*}[t]
\centering
\begin{tikzpicture}
  \matrix (m) [matrix of math nodes, row sep=3em,
    column sep=3em]{
    & F_N(t+1) & & S_N(t+1) \\
    F_N(t)& & S_N(t) & \\
    & F_R(t+1) & & S_R(t+1) \\
    F_R(t) & & S_R(t) & \\};
  \path[-stealth]
    (m-1-2) 
            edge [densely dotted] (m-3-2)
    (m-1-4) edge  node [right] {$d'$} (m-3-4)
            edge node [below] {$\phi$} (m-1-2) 
    (m-2-1) 
            edge (m-2-3) edge node [left] {$d$} (m-4-1)
            edge node [above] {$\zeta_N$} (m-1-2)
    (m-3-2) edge [densely dotted] (m-3-4)
            
    (m-4-1) edge node [below] {$\psi$} (m-4-3)
            edge node [densely dotted] [above] {$\zeta_R$}  (m-3-2)
    (m-2-3) edge [-,line width=6pt,draw=white] (m-4-3)
            edge (m-4-3)
            edge [-,line width=6pt,draw=white] (m-2-1)
            edge (m-2-1);
\end{tikzpicture}
\caption{Commutative diagram expressing $\zeta_R$ and $\zeta_N$. Some edge labels were removed for clarity. Refer to figure \ref{fig:retelling_1} on page \pageref{fig:retelling_1}.}
\label{fig:zeta_1}
\end{figure*}
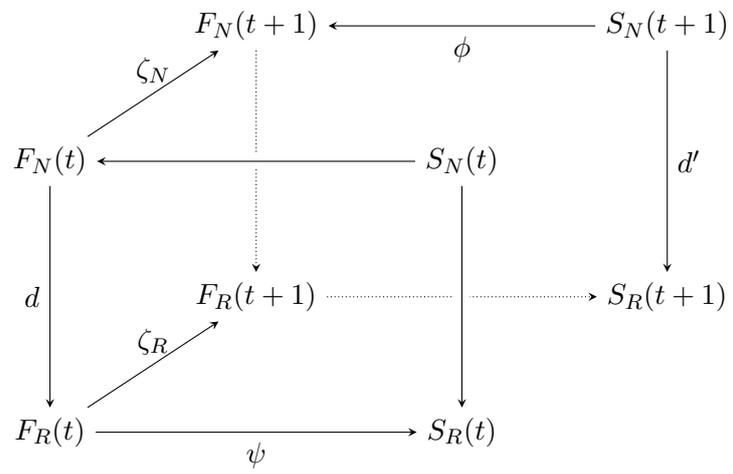
\end{document}